\documentclass[11pt,letterpaper]{article}
\usepackage{naaclhlt2016}
\usepackage{times}
\usepackage{latexsym}

\usepackage{mathtools}
\usepackage{url}
\usepackage{amsmath}
\usepackage{multirow}
\usepackage[normalem]{ulem}
\usepackage{color}
\usepackage[usenames,dvipsnames]{xcolor}
\usepackage{enumitem}
\usepackage[lined,boxed,commentsnumbered]{algorithm2e}
\usepackage{graphicx}
\usepackage{epstopdf}
\usepackage{subfig}
\usepackage{mdwlist}
\usepackage{enumitem}
\usepackage{amsfonts}
\usepackage{amsmath}
\usepackage{mathtools}
\usepackage{setspace}

\naaclfinalcopy 
\def\naaclpaperid{***} 

\title{Neural Network-Based Abstract Generation for Opinions and Arguments}

\author{Lu Wang \\
  College of Computer and Information Science\\
  Northeastern University \\
  Boston, MA 02115 \\
  {\tt luwang@ccs.neu.edu} \\\And
  Wang Ling \\
  Google DeepMind \\
  London, N1 0AE \\
  {\tt lingwang@google.com}
   \\}

\begin{document}

\maketitle

\begin{abstract}
We study the problem of generating abstractive summaries for opinionated text. We propose an attention-based neural network model that is able to absorb information from multiple text units to construct informative, concise, and fluent summaries. An importance-based sampling method is designed to allow the encoder to integrate information from an important subset of input. Automatic evaluation indicates that our system outperforms state-of-the-art abstractive and extractive summarization systems on two newly collected datasets of movie reviews and arguments. Our system summaries are also rated as more informative and grammatical in human evaluation.
\end{abstract}

\section{Introduction}
Collecting opinions from others is an integral part of our daily activities. Discovering what other people think can help us navigate through different aspects of life, ranging from making decisions on regular tasks to judging fundamental societal issues and forming personal ideology. To efficiently absorb the massive amount of opinionated information, there is a pressing need for automated systems that can generate concise and fluent opinion summary about an entity or a topic. 
In spite of substantial researches in opinion summarization, the most prominent approaches mainly rely on \textit{extractive summarization} methods, where phrases or sentences from the original documents are selected for inclusion in the summary~\cite{Hu:2004:MSC:1014052.1014073,Lerman:2009:SSE:1609067.1609124}. One of the problems that extractive methods suffer from is that they unavoidably include secondary or redundant information. On the contrary, \textit{abstractive summarization} methods, which are able to generate text beyond the original input, can produce more coherent and concise summaries.

\begin{figure}[t]
        {\fontsize{8.5}{9}\selectfont
        \begin{tabular}{|p{75mm}|}
        \hline
        \underline{\textbf{Movie}}: {\it The Martian} \\
        \textbf{Reviews}: \\
		- One the \textcolor{blue}{\bf smartest}, sweetest, and most satisfyingly suspenseful sci-fi films in years.\\
		- 
		...an intimate sci-fi epic that is \textcolor{blue}{\bf smart}, spectacular and stirring.\\
		- The Martian is a \textcolor{red}{\bf thrilling}, human and moving sci-fi picture that is easily the most emotionally engaging film Ridley Scott has made...\\
        - It's pretty sunny and often \textcolor{Plum}{\bf funny}, a space oddity for a director not known for pictures with \textcolor{Plum}{\bf a sense of humor}.\\
		- The Martian \textcolor{OliveGreen}{\bf highlights the book's best qualities}, tones down its worst, and adds its own style...\\ 
        \textbf{Opinion Consensus (Summary)}: \textcolor{blue}{\bf Smart}, \textcolor{red}{\bf thrilling}, and surprisingly \textcolor{Plum}{\bf funny}, The Martian offers \textcolor{OliveGreen}{\bf a faithful adaptation of the bestselling book} that brings out the best in leading man Matt Damon and director Ridley Scott.\\
        \hline
        
        \end{tabular}

        \begin{tabular}{|p{75mm}|}
        \hline
        \underline{\textbf{Topic}}: {\it This House supports the death penalty.} \\
        \textbf{Arguments}: \\
        - The state has a responsibility to protect the lives of innocent citizens, and enacting the death penalty may save lives by \textcolor{BurntOrange}{\bf reducing the rate of violent crime}.\\
        - While the prospect of life in prison may be frightening, surely death is a more daunting prospect.\\
        - A 1985 study by Stephen K. Layson at the University of North Carolina showed that a single execution \textcolor{BurntOrange}{\bf deters 18 murders}.\\
        - Reducing the wait time on death row prior to execution can dramatically increase its \textcolor{BurntOrange}{\bf deterrent effect} in the United States.\\
        \textbf{Claim (Summary)}: The death penalty \textcolor{BurntOrange}{\bf deters crime}.\\
        \hline
        
        \end{tabular}
        
        }
		\caption{\fontsize{9}{10}\selectfont Examples for an opinion consensus of professional reviews (critics) about movie ``{\it The Martian}'' from {\tt www.rottentomatoes.com}, and a claim about ``death penalty'' supported by arguments from {\tt idebate.org}. Content with similar meaning is highlighted in the same color.}
\label{fig:abstract_generation}
\end{figure}

In this paper, we present {\it an attention-based neural network model for generating abstractive summaries of opinionated text}. Our system takes as input a set of text units containing opinions about the same topic (e.g. reviews for a movie, or arguments for a controversial social issue), and then outputs a one-sentence abstractive summary that describes the opinion consensus of the input. 

Specifically, we investigate our abstract generation model on two types of opinionated text: \textit{movie reviews} and \textit{arguments on controversial topics}. Examples are displayed in Figure~\ref{fig:abstract_generation}. The first example contains a set of professional reviews (or critics) about movie ``The Martian'' and an opinion consensus written by an editor. It would be more useful to automatically generate fluent opinion consensus rather than simply extracting features (e.g. plot, music, etc) and opinion phrases as done in previous summarization work~\cite{Zhuang:2006:MRM:1183614.1183625,Li:2010:SRM:1873781.1873855}. 
%
The second example lists a set of arguments on ``death penalty'', where each argument supports the central claim ``death penalty deters crime''. Arguments, as a special type of opinionated text, contain reasons to persuade or inform people on certain issues. Given a set of arguments on the same topic, we aim at investigating the capability of our abstract generation system for the novel task of \textit{claim generation}. 

Existing abstract generation systems for opinionated text mostly take an approach that first identifies salient phrases, and then merges them into sentences~\cite{bing-EtAl:2015:ACL-IJCNLP,ganesan2010opinosis}. 
Those systems are not capable of generating new words, and the output summary may suffer from ungrammatical structure. 
Another line of work requires a large amount of human input to enforce summary quality. For example, \newcite{gerani-EtAl:2014:EMNLP2014} utilize a set of templates constructed by human, which are filled by extracted phrases to generate grammatical sentences that serve different discourse functions. 


To address the challenges above, we propose to use an attention-based abstract generation model --- a data-driven approach trained to generate informative, concise, and fluent opinion summaries. 
Our method is based on the recently proposed framework of neural encoder-decoder models~\cite{KalchbrennerB13,SutskeverVL14}, which translates a sentence in a source language into a target language. 
Different from previous work, our summarization system is designed to support multiple input text units. An attention-based model~\cite{DBLP:journals/corr/BahdanauCB14} is deployed to allow the encoder to automatically search for salient information within context. 
Furthermore, we propose an importance-based sampling method so that the encoder can integrate information from an important subset of input text. The importance score of a text unit is estimated from a novel regression model with pairwise preference-based regularizer. With importance-based sampling, our model can be trained within manageable time, and is still able to learn from diversified input. 

We demonstrate the effectiveness of our model on two newly collected datasets for movie reviews and arguments. 
Automatic evaluation by BLEU~\cite{papineni2002bleu} indicates that our system outperforms the state-of-the-art extract-based and abstract-based methods on both tasks. For example, we achieved a BLEU score of 24.88 on Rotten Tomatoes movie reviews, compared to 19.72 by an abstractive opinion summarization system from~\newcite{ganesan2010opinosis}. ROUGE evaluation~\cite{Lin:2003:AES:1073445.1073465} also indicates that our system summaries have reasonable information coverage. 
Human judges further rated our summaries to be more informative and grammatical than compared systems.

\section{Data Collection}
\label{sec:data}
We collected two datasets for movie reviews and arguments on controversial topics with gold-standard abstracts.\footnote{The datasets can be downloaded from \url{http://www.ccs.neu.edu/home/luwang/}.}
Rotten Tomatoes (\url{www.rottentomatoes.com}) is a movie review website that aggregates both professional critics and user-generated reviews (henceforth {\it RottenTomatoes}). For each movie, a one-sentence critic consensus is constructed by an editor to summarize the opinions in professional critics. We crawled 246,164 critics and their opinion consensus for 3,731 movies (i.e. around 66 reviews per movie on average). We select 2,458 movies for training, 536 movies for validation and 737 movies for testing. The opinion consensus is treated as the gold-standard summary.

We also collect an argumentation dataset from \url{idebate.org} (henceforth {\it Idebate}), which is a Wikipedia-style website for gathering pro and con arguments on controversial issues. The arguments under each debate (or topic) are organized into different ``for'' and ``against'' points. Each point contains a one-sentence central claim constructed by the editors to summarize the corresponding arguments, and is treated as the gold-standard. For instance, on a debate about ``death penalty'', one claim is ``the death penalty deters crime'' with an argument ``enacting the death penalty may save lives by reducing the rate of violent crime'' (Figure~\ref{fig:abstract_generation}). 
We crawled 676 debates with 2,259 claims. We treat each sentence as an argument, which results in 17,359 arguments in total. 450 debates are used for training, 67 debates for validation, and 150 debates for testing.

\section{The Neural Network-Based Abstract Generation Model}
\label{sec:model}
In this section, we first define our problem in Section~\ref{sec:problem}, followed by model description. 
In general, we utilize a Long Short-Term Memory network for generating abstracts (Section~\ref{sec:decoder}) from a latent representation computed by an attention-based encoder (Section~\ref{encoder}). The encoder is designed to search for relevant information from input to better inform the abstract generation process. We also discuss an importance-based sampling method to allow encoder to integrate information from an important subset of input (Sections~\ref{mul_inputs} and~\ref{sec:scoring}). Post-processing (Section~\ref{sec:postprocess}) is conducted to re-rank the generations and pick the best one as the final summary.

\subsection{Problem Formulation}
\label{sec:problem}
In summarization, the goal is to generate a summary $y$, composed by the sequence of words $y_1,...,|y|$. Unlike previous neural encoder-decoder approaches which decode from only one input, our input consists of an arbitrary number of reviews or arguments (henceforth \textit{text units} wherever there is no ambiguity), denoted as $x=\{x^{1},...,x^{M}\}$. 
Each text unit $x^{k}$ is composed by a sequence of words $x^{k}_{1},...,x^{k}_{|x^{k}|}$. 
Each word takes the form of a representation vector, which is initialized randomly or by pre-trained embeddings~\cite{journals/corr/abs-1301-3781}, and updated during training. 
The summarization task is defined as finding $\hat{y}$, which is the most likely sequence of words $\hat{y}_1,...,\hat{y}_N$ such that:

{\small
\begin{equation}
\hat{y} = argmax_{y} \log P(y|x)
\end{equation}
}
where $\log P(y|x)$ denotes the conditional log-likelihood of the output sequence $y$, given the input text units $x$. 
In the next sections, we describe the attention model used to model $\log P(y|x)$. 


\subsection{Decoder}
\label{sec:decoder}
Similar as previous work~\cite{DBLP:journals/corr/SutskeverVL14,DBLP:journals/corr/BahdanauCB14}, we decompose $\log P(y|x)$ into a sequence of word-level predictions:

{\small
\begin{equation}
\log P(y|x) = \sum_{j=1,...,|y|} \log P(y_j|y_1,...,y_{j-1}, x)
\end{equation}
}
where each word $y_j$ is predicted conditional on the previously generated $y_1,...,y_{j-1}$ and input $x$. 
The probability is estimated by standard word softmax:

{\small
\begin{equation}
p(y_j|y_1,...,y_{j-1}, x) = softmax(\mathbf{h}_j)
\end{equation}
}
$\mathbf{h}_j$ is the Recurrent Neural Networks (RNNs) state variable at timestamp $j$, which is modeled as:

{\small
\begin{equation}
\mathbf{h}_j = g(\mathbf{y}_{j-1},\mathbf{h}_{j-1},\mathbf{s})
\end{equation}
}
Here $g$ is a recurrent update function for generating the new state $\mathbf{h}_j$ from the representation of previously generated word $\mathbf{y}_{j-1}$ (obtained from a word lookup table), the previous state $\mathbf{h}_{j-1}$, and the input text representation $\mathbf{s}$ (see Section~\ref{encoder}). 

In this work, we implement $g$ using a Long Short-Term Memory (LSTM) network~\cite{Hochreiter:1997:LSM:1246443.1246450}, which has been shown to be effective at capturing long range dependencies. Here we summarize the update rules for LSTM cells, and refer readers to the original work~\cite{Hochreiter:1997:LSM:1246443.1246450} for more details. 
Given an arbitrary input vector $\mathbf{u}_{j}$ at timestamp $j-1$ and the previous state $\mathbf{h}_{j-1}$, a typical LSTM defines the following update rules:

{\small
\begin{align}
\label{lstm}
\begin{split}
\mathbf{i}_j &= \sigma(\mathbf{W}_{iu} \mathbf{u}_j + \mathbf{W}_{ih} \mathbf{h}_{j-1} + \mathbf{W}_{ic} \mathbf{c}_{j-1} + \mathbf{b}_i)
\\
\mathbf{f}_j &= \sigma(\mathbf{W}_{fu} \mathbf{u}_j + \mathbf{W}_{fh} \mathbf{h}_{j-1} + \mathbf{W}_{fc} \mathbf{c}_{j-1} + \mathbf{b}_f)
\\
\mathbf{c}_j &= \mathbf{f}_j \odot \mathbf{c}_{j-1} + \mathbf{i}_j \odot \tanh(\mathbf{W}_{cu}\mathbf{u}_j + \mathbf{W}_{ch}\mathbf{h}_{j-1} + \mathbf{b}_c)
\\
\mathbf{o}_j &= \sigma(\mathbf{W}_{ou} \mathbf{u}_j + \mathbf{W}_{oh} \mathbf{h}_{j-1} + \mathbf{W}_{oc} \mathbf{c}_{j} + \mathbf{b}_o)
\\
\mathbf{h}_j &= \mathbf{o}_j \odot \tanh(\mathbf{c}_j)
\end{split}
\end{align}
}
$\sigma$ is component-wise logistic sigmoid function, and $\odot$ denotes Hadamard product. Projection matrices $\mathbf{W}_{\ast \ast}$ and biases $\mathbf{b}_{\ast}$ are parameters to be learned during training.

Long range dependencies are captured by the cell memory $\mathbf{c}_j$, which is updated linearly to avoid the vanishing gradient problem. It is accomplished by predicting two vectors $\mathbf{i}_j$ and $\mathbf{f}_j$, which determine what to keep and what to forget from the current timestamp. Vector $\mathbf{o}_j$ then decides on what information from the new cell memory $\mathbf{c}_j$ can be passed to the new state $\mathbf{h}_j$. Finally, the model concatenates the representation of previous output word $\mathbf{y}_{j-1}$ and the input representation $\mathbf{s}$ (see Section~\ref{encoder}) as $\mathbf{u}_j$, which serves as the input at each timestamp. 

\subsection{Encoder}
\label{encoder}

The representation of input text units $\mathbf{s}$ is computed using an attention model~\cite{DBLP:journals/corr/BahdanauCB14}. Given a single text unit $x_1,...,x_{|x|}$ and the previous state $\mathbf{h_{j}}$, the model generates $\mathbf{s}$ as a weighted sum:

{\small
\begin{equation}
\label{attention}
\sum_{i = {1,...,|x|}} a_i \mathbf{b}_i
\end{equation}
}
where $a_i$ is the attention coefficient obtained for word $x_i$, and $\mathbf{b}_i$ is the context dependent representation of $x_i$. In our work, we construct $\mathbf{b}_i$ by building a bidirectional LSTM over the whole input sequence $x_1,...,x_{|x|}$ and then combining the forward and backward states. Formally, we use the LSTM formulation from Eq.~\ref{lstm} to generate the forward states $\mathbf{h}^f_1,...,\mathbf{h}^f_{|x|}$ by setting $\mathbf{u}_j=\mathbf{x}_j$ (the projection word $x_j$ using a word lookup table). Likewise, the backward states $\mathbf{h}^b_{|x|},...,\mathbf{h}^b_{1}$ are generated using a backward LSTM by feeding the input in the reverse order, that is, $\mathbf{u}_j=\mathbf{x}_{|x|-j+1}$. The coefficients $a_i$ are computed with a softmax over all input:

{\small
\begin{equation}
a_i = softmax(v(\mathbf{b}_i,\mathbf{h}_{j-1}))
\end{equation}
}
where function $v$ computes the affinity of each word $x_i$ and the current output context $\mathbf{h}_{j-1}$ --- how likely the input word is to be used to generate the next word in summary. We set {\small $v(\mathbf{b}_i,\mathbf{h}_{j-1})=\mathbf{W}_s \cdot \tanh(\mathbf{W}_{cg}\mathbf{b}_i + \mathbf{W}_{hg}\mathbf{h}_{j-1})$}, where $\mathbf{W}_{\ast}$ and $\mathbf{W}_{\ast \ast}$ are parameters to be learned.

\subsection{Attention Over Multiple Inputs}
\label{mul_inputs}
A key distinction between our model and existing sequence-to-sequence models~\cite{DBLP:journals/corr/SutskeverVL14,DBLP:journals/corr/BahdanauCB14} is that our input consists of multiple separate text units. 
Given an input of $N$ text units, i.e. $\{x^{k}_{1},...,x^{k}_{|x^{k}|}\}_{k=1}^{N}$, a simple extension would be to concatenate them into one sequence as $z=x^1_1,...,x^1_{|x^1|},\textsc{seg},x^2_1,...,x^2_{|x^2|}, \textsc{seg},x^{N}_1,...,x^{N}_{|x^N|}$, where \textsc{seg} is a special token that delimits inputs. 

However, there are two problems with this approach. Firstly, the model is sensitive to the order of text units. Moreover, $z$ may contain thousands of words. This will become a bottleneck for our model with a training time of $O(N|z|)$, since attention coefficients must be computed for all input words to generate each output word.


We address these two problems by sub-sampling from the input. The intuition is that even though the number of input text units is large, many of them are redundant or contain secondary information. As our task is to emphasize the main points made in the input, some of them can be removed without losing too much information. 
Therefore, we define an importance score $f(x^k)\in[0,1]$ for each document $x^k$ (see Section~\ref{sec:scoring}). During training, $K$ candidates are sampled from a multinomial distribution which is constructed by normalizing $f(x^k)$ for input text units. Notice that the training process goes over the training set multiple times, and our model is still able to learn from more than $K$ text units. 
For testing, top-$K$ candidates with the highest importance scores are collapsed in descending order into $z$.


\subsection{Importance Estimation}
\label{sec:scoring}
We now describe the importance estimation model, which outputs importance scores for text units. In general, we start with a ridge regression model, and add a regularizer to enforce the separation of summary-worthy text units from others. 

Given a cluster of text units $\{x^1,...,x^{M}\}$ and their summary $y$, we compute the number of overlapping content words between each text unit and summary $y$ as its gold-standard importance score. The scores are uniformly normalized to $[0,1]$. 
Each text unit $x^k$ is represented as an $d-$dimensional feature vector {\small $\mathbf{\mathbf{r_{k}}}\in \mathbb{R}^{d}$}, with label {\small $l_{k}$}. 
Text units in the training data are thus denoted with a feature matrix {\small $\mathbf{\tilde{R}}$} and a label vector {\small $\mathbf{\tilde{L}}$}. We aim at learning {\small $f(x^{k})=\mathbf{r_{k}}\cdot \mathbf{w}$} by minimizing 
{\small $||\mathbf{\tilde{R}} \mathbf{w} - \mathbf{\tilde{L}}||_{2}^{2} + \beta \cdot ||\mathbf{w}||_{2}^{2}$}. 
This is a standard formulation for ridge regression, and we use features in Table~\ref{tab:features_importance}.  
Furthermore, pairwise preference constraints have been utilized for learning ranking models~\cite{Joachims:2002:OSE:775047.775067}. We then consider adding a {\it pairwise preference-based regularizing constraint} to incorporate a bias towards summary-worthy text units:
{\small $\lambda\cdot \sum_{\mathcal{T}} \sum_{x^{p}, x^{q}\in \mathcal{T}, l_{p} > 0, l_{q} =0} ||(\mathbf{r_{p}}-\mathbf{r_{q}})\cdot \mathbf{w}- 1||_{2}^{2}$}, 
where $\mathcal{T}$ is a cluster of text units to be summarized. 
Term {\small $(\mathbf{r_{p}}-\mathbf{r_{q}})\cdot \mathbf{w}$} enforces the separation of summary-worthy text from the others. 
We further construct {\small$\mathbf{\tilde{R}^{\prime}}$} to contain all the pairwise differences {\small$(\mathbf{r_{p}}-\mathbf{r_{q}})$}. {\small$\mathbf{\tilde{L}^{\prime}}$} is a vector of the same size as {\small$\mathbf{\tilde{R}^{\prime}}$} with each element as $1$. The objective function becomes:

{\small
\begin{equation}
J(\mathbf{w})=||\mathbf{\tilde{R}} \mathbf{w} - \mathbf{\tilde{L}}||_{2}^{2} + \lambda\cdot ||\mathbf{\tilde{R}^{\prime}} \mathbf{w} - \mathbf{\tilde{L}^{\prime}}||_{2}^{2}+\beta\cdot ||\mathbf{w}||_{2}^{2}
\end{equation}
}
$\lambda$, $\beta$ are tuned on development set. With {\small $\tilde{\boldsymbol \beta}=\beta \cdot \mathbf{I_{d}}$} and {\small $\tilde{\boldsymbol \lambda} = \lambda \cdot \mathbf{I_{|R^{\prime}|}}$}, \textbf{closed-form} solution for {\small $\hat{\mathbf{w}}$} is:

{\small
\begin{equation}
\hat{\mathbf{w}}=(\mathbf{\tilde{R}^{T}\tilde{R}+\tilde{R}^{\prime T} \tilde{\boldsymbol \lambda} \tilde{R}^{\prime}}+\tilde{\boldsymbol \beta})^{-1}(\mathbf{\tilde{R}^{T}\tilde{L}+\tilde{R}^{\prime T} \tilde{\boldsymbol \lambda} \tilde{L}^{\prime}})
\end{equation}
}

\begin{table}[htbp]
\centering
    {\fontsize{9}{10}\selectfont
    \setlength{\baselineskip}{0pt}
    \setlength{\tabcolsep}{0.8mm}
    \begin{tabular}{|l|l|}
    \hline

	- num of words & - category in General Inquirer\\
	- unigram & ~\cite{stone66}\\
	- num of POS tags & - num of positive/negative/neutral \\ 
	- num of named entities & ~ words (General Inquirer,\\
	- centroidness~\cite{Radev01experimentsin} & ~MPQA~\cite{Wilson:2005:RCP})\\
	- avg/max TF-IDF scores & \\ 
	\hline
	\end{tabular}
	}
    \caption{\fontsize{9}{10}\selectfont Features used for text unit importance estimation.} 
	\label{tab:features_importance}
\end{table}

\subsection{Post-processing}
\label{sec:postprocess}
For testing phase, we re-rank the $n$-best summaries according to their cosine similarity with the input text units. The one with the highest similarity is included in the final summary. 
Uses of more sophisticated re-ranking methods~\cite{Charniak:2005:CNB:1219840.1219862,konstas-lapata:2012:ACL2012} will be investigated in future work.

\section{Experimental Setup}
\label{sec:setup}
\paragraph{Data Pre-processing.}
We pre-process the datasets with Stanford CoreNLP~\cite{manning-etal:2014:ACLDemo} for tokenization and extracting POS tags and dependency relations. For RottenTomatoes dataset, we replace movie titles with a generic label in training, and substitute it with the movie name if there is any generic label generated in testing.

\paragraph{Pre-trained Embeddings and Features.}
The size of word representation is set to 300, both for input and output words. These can be initialized randomly or using pre-trained embeddings learned from Google news~\cite{journals/corr/abs-1301-3781}. 
We also extend our model with additional features described in Table~\ref{tab:features_model}. Discrete features, such as POS tags, are mapped into word representation via lookup tables. For continuous features (e.g TF-IDF scores), they are attached to word vectors as additional values. 

\begin{table}[ht]
\centering
    {\fontsize{9}{10}\selectfont
    \setlength{\baselineskip}{0pt}
    \setlength{\tabcolsep}{0.8mm}
    \begin{tabular}{|l|l|}
    \hline

	- part of a named entity? & - category in General Inquirer\\
	- capitalized? & - sentiment polarity\\
	- POS tag &  ~ (General Inquirer, MPQA)\\
	- dependency relation & - TF-IDF score \\
	\hline
	\end{tabular}
	}
    \caption{\fontsize{9}{10}\selectfont Token-level features used for abstract generation.} 
	\label{tab:features_model}
\end{table}

\paragraph{Hyper-parameters and Stop Criterion.}
The LSTMs (Equation~\ref{lstm}) for the decoder and encoders are defined with states and cells of 150 dimensions. The attention of each input word and state pair is computed by being projected into a vector of 100 dimensions (Equation~\ref{attention}). 

Training is performed via Adagrad~\cite{Duchi:2011:ASM:1953048.2021068}. It terminates when performance does not improve on the development set. We use BLEU (up to 4-grams)~\cite{papineni2002bleu} as evaluation metric, which computes the precision of n-grams in generated summaries with gold-standard abstracts as the reference. 
Finally, the importance-based sampling rate ($K$) is set to 5 for experiments in Sections~\ref{sec:autoeval} and~\ref{sec:humaneval}.

Decoding is performed by beam search with a beam size of 20, i.e. we keep 20 most probable output sequences in stack at each step. 
Outputs with \texttt{end of sentence} token are also considered for re-ranking. Decoding stops when every beam in stack generates the \texttt{end of sentence} token.




\section{Results}
\label{sec:result}
\subsection{Importance Estimation Evaluation}
We first evaluate the importance estimation component described in Section~\ref{sec:scoring}. We compare with Support Vector Regression (SVR)~\cite{smola1997support} and two baselines: (1) a \textit{length baseline} that ranks text units based on their length, and (2) a \textit{centroid baseline} that ranks text units according to their centroidness, which is computed as the cosine similarity between a text unit and centroid of the cluster to be summarized~\cite{Erkan:2004:LGL:1622487.1622501}.

\begin{figure}[thbp]
\subfloat
{
        \hspace{-4mm}
    \includegraphics[width=42mm,height=35mm]{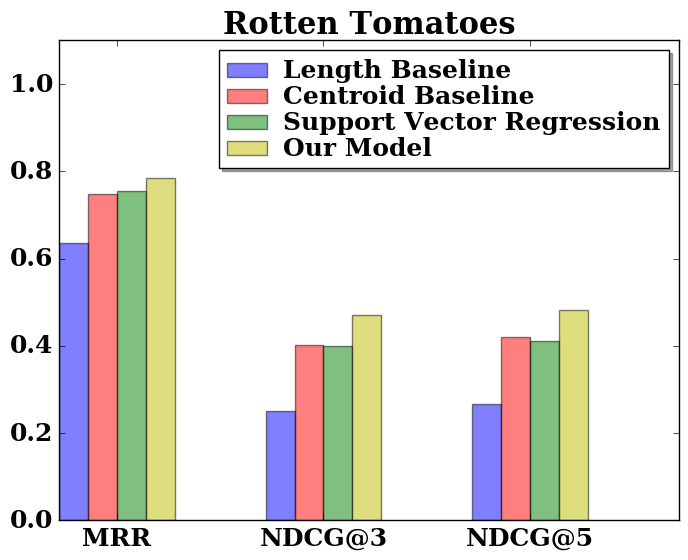}
}
\subfloat
{       \hspace{-3mm}
    \includegraphics[width=42mm,height=35mm]{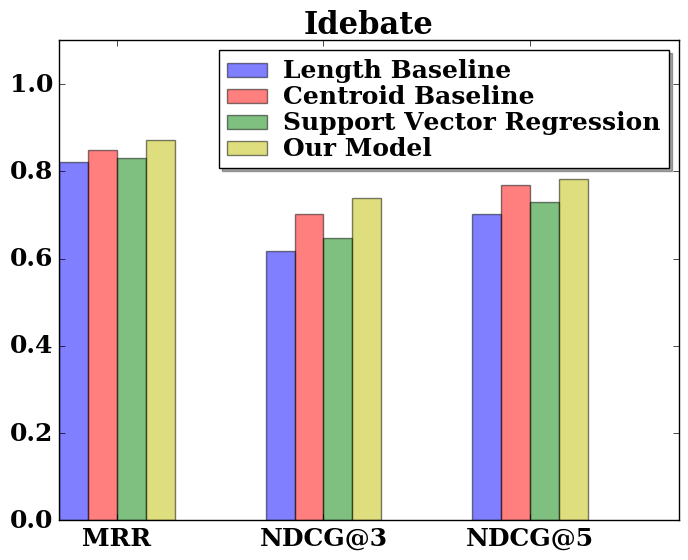}
}
\caption{\fontsize{9}{10}\selectfont Evaluation of importance estimation by mean reciprocal rank (MRR), and normalized discounted cumulative gain at top 3 and 5 returned results (NDCG@3 and NDCG@5). Our regression model with pairwise preference-based regularizer uniformly outperforms baseline systems on both datasets.}
\label{fig:result_ranking}
\end{figure}

We evaluate using mean reciprocal rank (MRR), and normalized discounted cumulative gain at top 3 and 5 returned results (NDCG@3). Text units are considered relevant if they have at least one overlapping content word with the gold-standard summary. 
From Figure~\ref{fig:result_ranking}, we can see that our importance estimation model produces uniformly better ranking performance on both datasets.

\begin{table*}[thbp]
    {\fontsize{9}{9}\selectfont
    \setlength{\baselineskip}{0pt}
    \centering
    \begin{tabular}{|l|l l l l|l l l l|}
    \hline
    &\multicolumn{4}{|c|}{\textit{RottenTomatoes}} &\multicolumn{4}{|c|}{\textit{Idebate}}\\    
	& \textbf{Length} & \textbf{BLEU} & \textbf{METEOR} & \textbf{ROUGE} & \textbf{Length} & \textbf{BLEU} & \textbf{METEOR} & \textbf{ROUGE}\\
	
	\textbf{Extract-Based Systems} &  &  &  &  &  &  &  &\\
	\textsc{Longest} & 47.9 & 8.25 & {\bf 8.43} & {\bf 6.43}					& 44.0 & 6.36 & 10.22  & 12.65\\
	\textsc{LexRank} & 16.7 & 19.93 & 5.59 & 3.98						& 26.5 & 13.39 & 9.33  & 10.58\\
	\textsc{Submodular} & 16.8 & 17.22 & 4.89 & 3.01						& 23.2 & 15.09 & {\bf 10.76}  & {\bf 13.67}\\
	
	\textbf{Abstract-Based Systems} &  &  &  &  &  &  & & \\
	\textsc{Opinosis} & 22.0 & 19.72 & 6.07 & 4.90						& -- & -- & -- & --\\
	
	{\textsc{Our Systems}} &  &  &  &  &  &  & &\\
	\textit{~words} & 15.7 & 19.88 & 6.07 &	5.05											& 14.4 & 22.55$^{\ast}$ & 7.38 & 8.37\\
	\textit{~words (pre-trained)} & 15.8 & 23.22$^{\ast}$ & {\it 6.51}  & {\it 5.70}					& 13.9 & 23.93$^{\ast}$ & 7.42 & 9.09\\
	\textit{~words + features} & 17.5 & 19.73 &  6.43  & 5.53									& 13.5 & 23.65$^{\ast}$ & 7.33 & 7.79\\
	\textit{~words (pre-trained) + features } & 14.2 & {\bf 24.88}$^{\ast}$ & 6.00  & 4.96		& 13.0 & {\bf 25.84}$^{\ast}$ & 7.56 & 8.81\\

	\hline
	\end{tabular}

	}
    \caption{\fontsize{9}{10}\selectfont Automatic evaluation results by BLEU, METEOR, and ROUGE SU-4 scores (multiplied by 100) for abstract generation systems. The average lengths for human written summaries are $11.5$ and $24.6$ for RottenTomatoes and Idebate. The best performing system for each column is highlighted in {\bf boldface}, where our system with pre-trained word embeddings and additional features achieves the best BLEU scores on both datasets. Our systems that are statistically significantly better than the comparisons are highlighted with $\ast$ (two-tailed Wilcoxon signed rank test, $p<0.05$). 
 	Our system also has the best METEOR and ROUGE scores (in {\it italics}) on RottenTomatoes dataset among learning-based systems.}
	\label{tab:result_main}
\end{table*}

\subsection{Automatic Summary Evaluation}
\label{sec:autoeval}

For automatic summary evaluation, we consider three popular metrics. ROUGE~\cite{Lin:2003:AES:1073445.1073465} is employed to evaluate n-grams recall of the summaries with gold-standard abstracts as reference. ROUGE-SU4 (measures unigram and skip-bigrams separated by up to four words) is reported. 
We also utilize BLEU, a precision-based metric, which has been used to evaluate various language generation systems~\cite{chiang2005hierarchical,angeli2010simple,karpathy2014deep}. 
We further consider METEOR~\cite{denkowski:lavie:meteor-wmt:2014}. As a recall-oriented metric, it calculates similarity between generations and references by considering synonyms and paraphrases. 

For comparisons, we first compare with an abstractive summarization method presented in~\newcite{ganesan2010opinosis} on the RottenTomatoes dataset. \newcite{ganesan2010opinosis} utilize a graph-based algorithm to remove repetitive information, and merge opinionated expressions based on syntactic structures of product reviews.\footnote{We do not run this model on Idebate because it relies on high redundancy to detect repetitive expressions, which is not observed on Idebate.}
For both datasets, we consider two extractive summarization approaches: (1) \textsc{LexRank}~\cite{Erkan:2004:LGL:1622487.1622501} is an unsupervised method that computes text centrality based on PageRank algorithm; (2) \newcite{Sipos:2012:LLS:2380816.2380846} propose a supervised \textsc{submodular} summarization model which is trained with Support Vector Machines. 
In addition, \textsc{longest} sentence is picked up as a baseline.

Four variations of our system are tested. One uses randomly initialized word embeddings. The rest of them use pre-trained word embeddings, additional features in Table~\ref{tab:features_model}, and their combination. For all systems, we generate a one-sentence summary.




Results are displayed in Table~\ref{tab:result_main}. Our system with pre-trained word embeddings and additional features achieves the best BLEU scores on both datasets (in {\bf boldface}) with statistical significance (two-tailed Wilcoxon signed rank test, $p<0.05$). Notice that our system summaries are conciser (i.e. shorter on average), which lead to higher scores on precision based-metrics, e.g. BLEU, and lower scores on recall-based metrics, e.g. METEOR and ROUGE. 
On RottenTomatoes dataset, where summaries generated by different systems are similar in length, our system still outperforms other methods in METEOR and ROUGE in addition to their significantly better BLEU scores. 
This is not true on Idebate, since the length of summaries by extract-based systems is significantly longer. But the BLEU scores of our system are considerably higher. 
Among our four systems, models with pre-trained word embeddings in general achieve better scores. Though additional features do not always improve the performance, we find that they help our systems converge faster.

\subsection{Human Evaluation on Summary Quality}
\label{sec:humaneval}

\begin{table}[htbp]
    {\fontsize{9}{10}\selectfont
   	\setlength{\tabcolsep}{0.7mm}
    
    \begin{tabular}{|l|c c c|c c|}
    \hline
	& \textbf{Info} & \textbf{Gram} & \textbf{Comp} & \textbf{Avg Rank} & \textbf{Best\%}\\
	
	\textsc{LexRank} 		& 3.4 & 4.5 & 4.3 & 2.7 & 11.5\% \\
	\textsc{Opinosis} 		& 2.8 & 3.1 & 3.3 & 3.5 & 5.0\% \\
	\textsc{Our System} 	& {\bf 3.6} & {\bf 4.8} & 4.2 & {\bf 2.3} & {\bf 18.0\%} \\
	\hline
	\hline
	\textsc{Human Abstract} 			& 4.2 & 4.8 & 4.5 & 1.5 & 65.5\% \\
	\hline
	\end{tabular}

	}
    \caption{\fontsize{9}{10}\selectfont  Human evaluation results for abstract generation systems. Inter-rater agreement for overall ranking is 0.71 by Krippendorff's $\alpha$. Informativeness (\textbf{Info}), grammaticality (\textbf{Gram}), and Compactness (\textbf{Comp}) are rated on a 1 to 5 scale, with 5 as the best. Our system achieves the best informativeness and grammaticality scores among the three learning-based systems. Our summaries are ranked as the best in 18\% of the evaluations, and are also ranked higher than compared systems on average.}
	\label{tab:result_human}
\end{table}

\begin{figure}[htbp]
        {\fontsize{9}{10}\selectfont
        \setlength{\tabcolsep}{0.6mm}
        \begin{tabular}{|p{78mm}|}
        \hline
        \underline{\textbf{Movie}}: {\it The Neverending Story} \\
        \textbf{Reviews}: (1) Here is a little adventure that fed on our uncultivated need to think, and wonder... (2) Magical storytelling targeted at children still fascinates. (3)...the art direction involved a lot of imagination.\\
        \textbf{Human}: A magical journey about the power of a young boy's imagination to save a dying fantasy land, The Neverending Story remains a much-loved kids adventure.\\
        \textbf{LexRank}: It pokes along at times and lapses occasionally into dark moments of preachy philosophy, but this is still a charming, amusing and harmless film for kids.\\
        \textbf{Opinosis}: The Neverending Story is a silly fantasy movie that often shows its age .\\
        \textbf{Our System}: The Neverending Story is an entertaining children's adventure, with heart and imagination to spare.\\
        \hline
        
        \underline{\textbf{Movie}}: {\it Joe Strummer: The Future is Unwritten} \\
        \textbf{Reviews}: (1) The late punk rock legend Joe Strummer is rendered fully human in Julian Temple's engrossing and all-encompassing portrait. (2) The movie fascinates not so much because of Strummer... but because of the way Temple organized and edited the film. (3) One of the most compelling documentary portraits of a musician yet made.\\
        \textbf{Human}: Displaying Joe Strummer warts and all, The Future is Unwritten succeeds as both an engrossing documentary and a comprehensive examination of one of music's most legendary figures.\\
        \textbf{LexRank}: Joe Strummer: The Future Is Unwritten is a film for fans -- really big fans .\\
        \textbf{Opinosis}: Joe Strummer: The Future Is Unwritten is for fans -- really big fans .\\
        \textbf{Our System}: Fascinating and insightful, Joe Strummer: The Future Is Unwritten is a thoroughly engrossing documentary.\\
        \hline
        \end{tabular}
        
        \begin{tabular}{|p{78mm}|}
        \hline
        \underline{\textbf{Topic}}: {\it This House would detain terror suspects without trial.} \\
        \textbf{Arguments}: (1) Governments must have powers to protect their citizens against threats to the life of the nation.(2) Everyone would recognise that rules that are applied in peacetime may not be appropriate during wartime. \\
        \textbf{Human}: Governments must have powers to protect citizens from harm.\\
        \textbf{LexRank}: This is not merely to directly protect citizens from political violence, but also because political violence handicaps the process of reconstruction in nation-building efforts.\\
        \textbf{Our System}: Governments have the obligation to protect citizens from harmful substances.\\
        \hline
        
        \underline{\textbf{Topic}}: {\it This House would replace Christmas with a festival for everyone.} \\
        \textbf{Arguments}: (1) Christmas celebrations in the Western world... do not respect the rights of those who are not religious. (2) States should instead be sponsoring and celebrating events that everyone can join in equally, regardless of religion, race or class.\\
        \textbf{Human}: States should respect the freedom from religion, as well as the freedom of religion.\\
        \textbf{LexRank}: For school children who do not share the majority-Christian faith, Christmas celebrations require either their participation when they do not want to, through coercion, or their non-participation and therefore isolation whilst everyone else celebrations their inclusiveness.\\
        \textbf{Our System}: People have a right to freedom of religion.\\
        \hline
        \end{tabular}
        }
		\caption{\fontsize{9}{10}\selectfont Sample summaries generated by different systems on movie reviews and arguments. We only show a subset of reviews and arguments due to limited space.}
\label{fig:sample_summary}
\end{figure}

For human evaluation, we consider three aspects: \textit{informativeness} that indicates how much salient information is contained in the summary, \textit{grammaticality} that measures whether a summary is grammatical, and \textit{compactness} that denotes whether a summary contains unnecessary information. Each aspect is rated on a 1 to 5 scale (5 is the best). The judges are also asked to give a ranking on all summary variations according to their overall quality.

We randomly sampled 40 movies from RottenTomatoes test set, each of which was evaluated by 5 distinct human judges. 
We hired 10 proficient English speakers for evaluation. Three system summaries (LexRank, Opinosis, and our system) and human-written abstract along with 20 representative reviews were displayed for each movie. Reviews with the highest gold-standard importance scores were selected.

Results are reported in Table~\ref{tab:result_human}. As it can be seen, our system outperforms the abstract-based system \textsc{Opinosis} in all aspects, and also achieves better informativeness and grammaticality scores than \textsc{LexRank}, which extracts sentences in their original form. 
Our system summaries are ranked as the best in 18\% of the evaluations, and has an average ranking of 2.3, which is higher than both \textsc{Opinosis} and \textsc{LexRank} on average. An inter-rater agreement of Krippendorff's $\alpha$ of 0.71 is achieved for overall ranking. 
This implies that our attention-based abstract generation model can produce summaries of better quality than existing summarization systems. We also find that our system summaries are constructed in a style closer to human abstracts than others. Sample summaries are displayed in Figure~\ref{fig:sample_summary}.

\subsection{Sampling Effect}

\begin{figure}[thbp]
\subfloat
{
        \hspace{-4mm}
    \includegraphics[width=42mm,height=35mm]{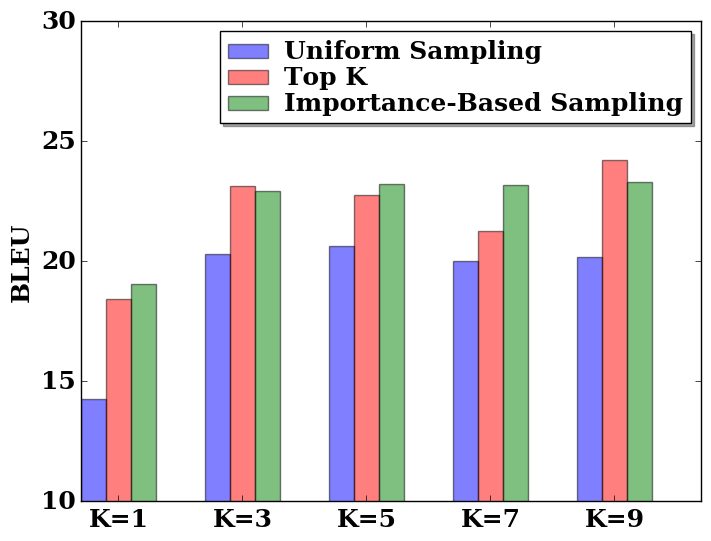}
}
\subfloat
{       \hspace{-3mm}
    \includegraphics[width=42mm,height=35mm]{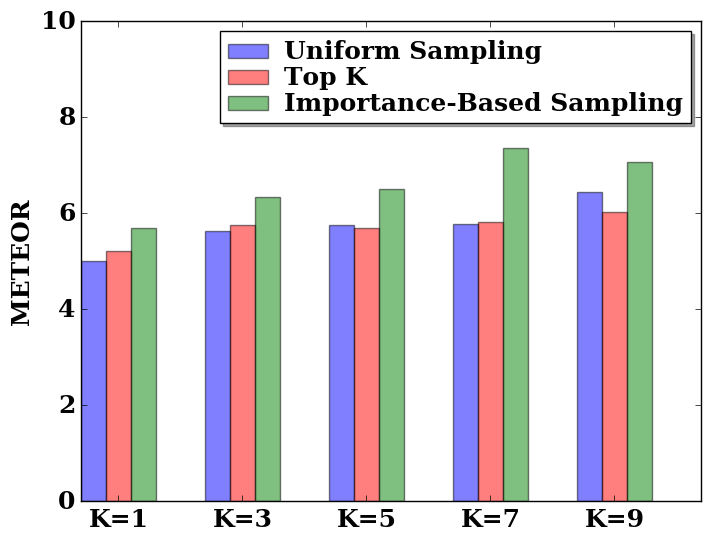}
}
\caption{\fontsize{9}{10}\selectfont Sampling effect on RottenTomatoes. }
\label{fig:sample_effect}
\end{figure}

We further investigate whether taking inputs sampled from distributions estimated by importance scores trains models with better performance than the ones learned from fixed input or uniformly-sampled input.  
Recall that we sample $K$ text units based on their importance scores (\textit{Importance-Based Sampling}). Here we consider two other setups: one is sampling $K$ text units uniformly from the input (\textit{Uniform Sampling}), another is picking $K$ text units with the highest scores (\textit{Top K}). We try various $K$ values. Results in Figure~\ref{fig:sample_effect} demonstrates that Importance-Based Sampling can produce comparable BLEU scores to Top K methods, while both of them outperform Uniform Sampling. For METEOR score, Importance-Based Sampling uniformly outperforms the other two methods\footnote{We observe similar results on the Idebate dataset}. 

\subsection{Further Discussion}
Finally, we discuss some other observations and potential improvements. 
%
First, applying the re-ranking component after the model generates $n$-best abstracts leads to better performance. Preliminary experiments show that simply picking the top-1 generations produces inferior results than re-ranking them with simple heuristics. This suggests that the current models are oblivious to some task specific issues, such as informativeness. Post-processing is needed to make better use of the summary candidates. For example, future work can study other sophisticated re-ranking algorithms~\cite{Charniak:2005:CNB:1219840.1219862,konstas-lapata:2012:ACL2012}.

Furthermore, we also look at the difficult cases where our summaries are evaluated to have lower informativeness. They are often much shorter than the gold-standard human abstracts, thus the information coverage is limited. In other cases, some generations contain incorrect information on domain-dependent facts, e.g. named entities, numbers, etc. For instance, a summary ``a poignant coming-of-age tale marked by a breakout lead performance from Cate Shortland'' is generated for movie ``Lore''. This summary contains ``Cate Shortland'' which is the director of the movie instead of actor. It would require semantic features to handle this issue, which has yet to be attempted.


\section{Related Work}
\label{sec:related}
Our work belongs to the area of opinion summarization. 
Constructing fluent natural language opinion summaries has mainly considered product reviews~\cite{Hu:2004:MSC:1014052.1014073,Lerman:2009:SSE:1609067.1609124}, community question answering~\cite{wang-EtAl:2014:Coling2}, and editorials~\cite{Paul:2010:SCV:1870658.1870665}. Extractive summarization approaches are employed to identify summary-worthy sentences. For example, \newcite{Hu:2004:MSC:1014052.1014073} first identify the frequent product features and then attach extracted opinion sentences to the corresponding feature. 
Our model instead utilizes abstract generation techniques to construct natural language summaries. As far as we know, we are also the first to study claim generation for arguments.

Recently, there has been a growing interest in generating abstractive summaries for news articles~\cite{bing-EtAl:2015:ACL-IJCNLP}, spoken meetings~\cite{wang-cardie:2013:ACL2013}, and product reviews~\cite{ganesan2010opinosis,di2014hybrid,gerani-EtAl:2014:EMNLP2014}. Most approaches are based on phrase extraction, from which an algorithm concatenates them into sentences~\cite{bing-EtAl:2015:ACL-IJCNLP,ganesan2010opinosis}. 
Nevertheless, the output summaries are not guaranteed to be grammatical. 
\newcite{gerani-EtAl:2014:EMNLP2014} then design a set of manually-constructed realization templates for producing grammatical sentences that serve different discourse functions. 
Our approach does not require any human-annotated rules, and can be applied in various domains. 

Our task is closely related to recent advances in neural machine translation~\cite{KalchbrennerB13,SutskeverVL14}. 
%
Based on the sequence-to-sequence paradigm, RNNs-based models have been investigated for compression~\cite{filippova-EtAl:2015:EMNLP} and summarization~\cite{filippova-EtAl:2015:EMNLP,rush-chopra-weston:2015:EMNLP,DBLP:journals/corr/HermannKGEKSB15} at sentence-level. 
Built on the attention-based translation model in~\newcite{DBLP:journals/corr/BahdanauCB14}, \newcite{rush-chopra-weston:2015:EMNLP} study the problem of constructing abstract for a single sentence. 
%
Our task differs from the models presented above in that our model carries out abstractive decoding from multiple sentences instead of a single sentence.

\section{Conclusion}
\label{sec:conclusion}
In this work, we presented a neural approach to generate abstractive summaries for opinionated text. 
We employed an attention-based method that finds salient information from different input text units to generate an informative and concise summary. To cope with the large number of input text, we deploy an importance-based sampling mechanism for model training. Experiments showed that our system obtained state-of-the-art results using both automatic evaluation and human evaluation.

\bibliographystyle{naaclhlt2016}

\end{document}